\documentclass[nohyperref, accepted]{article}

\usepackage{microtype}
\usepackage{graphicx}
\usepackage{subfigure}
\usepackage{booktabs} % for professional tables

\usepackage{hyperref}

\usepackage{icml2022}

\usepackage{amsmath}
\usepackage{amssymb}
\usepackage{mathtools}
\usepackage{amsthm}

\usepackage[capitalize,noabbrev]{cleveref}

\theoremstyle{plain}

\theoremstyle{definition}

\theoremstyle{remark}

\usepackage[textsize=tiny]{todonotes}

\icmltitlerunning{NERDA-Con: Extending NER models for Continual Learning}

\begin{document}

\twocolumn[
\icmltitle{NERDA-Con: Extending NER models for Continual Learning --- \\ Integrating Distinct Tasks and Updating Distribution Shifts}



\icmlsetsymbol{equal}{*}

\begin{icmlauthorlist}
\icmlauthor{Supriti Vijay}{equal,yyy,manipal}
\icmlauthor{Aman Priyanshu}{equal,comp,manipal}
\end{icmlauthorlist}

\icmlaffiliation{yyy}{Department of Computer Science}
\icmlaffiliation{comp}{Department of Information \& Communication Technology}
\icmlaffiliation{manipal}{Manipal Institute of Technology, Karnataka, India}

\icmlcorrespondingauthor{Supriti Vijay}{supriti.vijay@gmail.com }
\icmlcorrespondingauthor{Aman Priyanshu}{amanpriyanshusms2001@gmail.com}

\icmlkeywords{Machine Learning, ICML}

\vskip 0.3in
]



\printAffiliationsAndNotice{\icmlEqualContribution} % otherwise use the standard text.

\begin{abstract}

With increasing applications in areas such as biomedical information extraction pipelines and social media analytics, Named Entity Recognition (NER) has become an indispensable tool for knowledge extraction. However, with the gradual shift in language structure and vocabulary, NERs are plagued with distribution shifts, making them redundant or not as profitable without re-training. Re-training NERs based on Large Language Models (LLMs) from scratch over newly acquired data poses economic disadvantages. In contrast, re-training only with newly acquired data will result in Catastrophic Forgetting of previously acquired knowledge. Therefore, we propose NERDA-Con, a pipeline for training NERs with LLM bases by incorporating the concept of Elastic Weight Consolidation (EWC) into the NER fine-tuning NERDA pipeline \cite{nerda}. As we believe our work has implications to be utilized in the pipeline of continual learning and NER, we open-source our code as well as provide the fine-tuning library of the same name NERDA-Con at~\url{https://github.com/SupritiVijay/NERDA-Con} and~\url{https://pypi.org/project/NERDA-Con/}. 
\end{abstract}

\section{Introduction}

Named Entity Recognition (NER), an essential aspect of natural language processing, intends to identify spans of text corresponding to named entities and classify these from a set of pre-defined entity classes such as person (PER), location (LOC) or organization (ORG). With varied and 

\begin{figure}[ht]
% \vskip 0.2in
\begin{center}
\centerline{\includegraphics[width=\columnwidth,height=5cm]{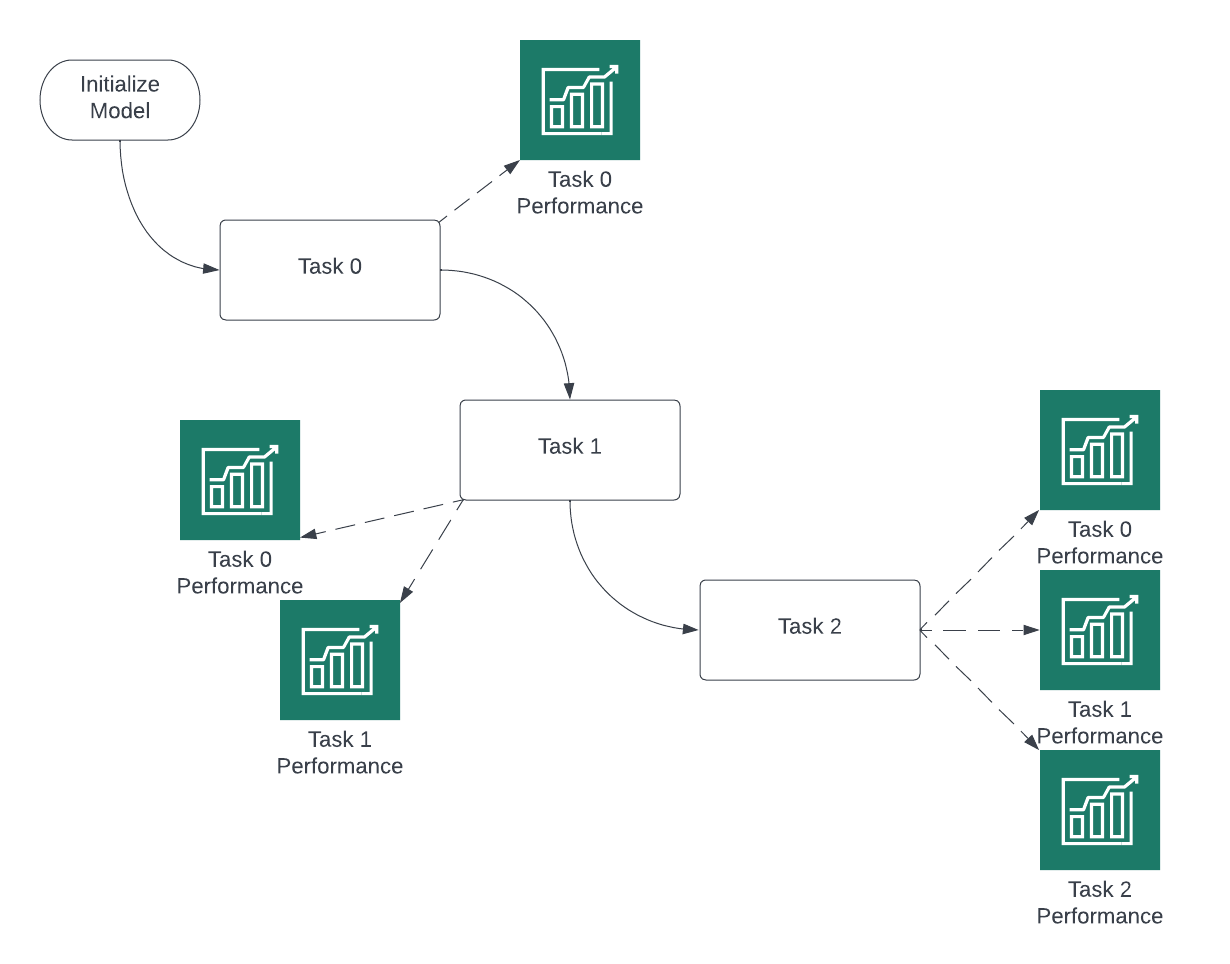}}
\caption{Overall framework for Continual Named Entity Recognition.}
\label{continual}
\end{center}
\vskip -0.3in
\end{figure}
 extensive applications spread across the fields of biomedical data to legal entity extractions, NERs have become an integral segment of automated knowledge extraction systems. However, Large Language Models (LLMs), which form the base of powerful NERs, are trained on a vast quantity of data to accomplish a said job. The assumption being that the model will meet similar data throughout testing. Unfortunately, real-world circumstances do not meet this need. Non-stationary processes cause slow or abrupt drifts in the data distribution \cite{gama2014survey, ditzler2015learning}, and the models must be constantly adjusted to account for the new changes. Continual learning offers an augmentable solution to the aforementioned problem.

To address this, we propose NERDA-Con, which aims to integrate continual learning with NER. We present the overall progression of continual NER in Figure~\ref{continual}. By incorporating EWC within NERDA's \cite{nerda} training module, we streamline the process of training over multiple tasks. We provide two variations to the algorithm, allowing users to encompass both (1) Distribution Shifts as well as (2) Separate Tasks.

An experimental study is then performed to verify the nature of knowledge retained across tasks. Our objectives of achieving improved performance over distribution shifts and differing tasks are formulated and described in detail in Section~\ref{methodology}. Unlike existing frameworks, NERDA-Con exploits Elastic Weight Consolidation (EWC) to regularize losses in future training calls. Our contributions to this manuscript can be summarized as follows:

\begin{enumerate}
    \item We reconsider training individual NERs on separate tasks by focussing its training on an exclusive task with intermediate sub-tasks pertaining to each data-node instead. We incorporate the concept of continual learning; specifically, EWC for training through said procedures.
    \item Demonstrate two experimental setting with the objectives of achieving high performance for (a) Distribution Shifts, \& (b) Separate Tasks. The experimental results on both objective benchmarks demonstrate that our framework outperforms the base/naive approaches.
    \item We open source a python library (NERDA-Con) for the construction of Continual Learning based NER models that employ LLMs directly integrated with the HuggingFace API. Allowing easy training, validation, and deployment of said models.
\end{enumerate}

\section{Related Work}

\subsection{Named Entity Recognition}
Research on NER, a prevailing method in NLP started with the use of conventional methods \cite{zhou-su-2002-named,chieu-ng-2002-named, 10.3115/1119176.1119196, 10.5555/1567594.1567618} which relied on handcraft features to build NER models. The nature of hand-crafted features made them susceptible to overfitting and difficult to transfer between languages. This led to the introduction of supervised learning techniques like Hidden Markov Models (HMM), Decision Trees, Maximum Entropy Models (ME), Support Vector Machines (SVM), Conditional Random Fields (CRF), which were trained on large annotated corpora to produce state of the art results for NER \cite{https://doi.org/10.48550/arxiv.2101.11420}. Due to the limit to structured text, semi-supervised learning methods were then utilized for a growing volume of unstructured text ie. webpages, to derive contextual information in an unsupervised manner\cite{10.1007/11766247_23, 10.3115/1072228.1072382, 10.5555/315149.315364,cucchiarelli-velardi-2001-unsupervised,10.5555/1597348.1597411}.

Recently, deep learning methods came to light since they significantly showed progress by automatically extracting high-level features and performing sequence tagging with neural networks \cite{https://doi.org/10.48550/arxiv.1505.05008, https://doi.org/10.48550/arxiv.1511.08308, https://doi.org/10.48550/arxiv.1603.01360, https://doi.org/10.48550/arxiv.1910.11470}. The rise of transformers in NLP brought promising results for, \cite{https://doi.org/10.48550/arxiv.1911.04474} proposed TENER, a NER architecture adopting an adapted Transformer Encoder to model the character-level features and word-level features. 

Thus, utilizing the best technique for NER, in our proposed algorithm, we use the NERDA Framework, an open-sourced tool that fine-tunes transformers for NER tasks for any arbitrary language to identify sensitive information from text.

\subsection{Continual Learning}

Continual learning refers to the ability to continually learn over time by accommodating new knowledge while retaining previously learned experiences \cite{PARISI201954}. Joint learning is an implementation accommodating the procedures of continual learning, which requires interleaving samples from each task \cite{Caruana1997}. However, this methodology requires memory from previous tasks and data samples, making it more resource constraining. This led to the development of Learning without Forgetting ($LwF$) \cite{li2017learning}, which only required samples form the task-at-hand for learning. It uses regularisation to compensate for the network forgetting an entire sequence of old data. However, this methodology still does not account for gradient flow and may restrain the training over the current task. Therefore, the restriction of only deviating neurons was proposed in "Overcoming catastrophic forgetting in neural networks" paper, which discusses the integration of the Fisher information matrix as a regularizer for the loss function \cite{ewc_paper}. We provide said equation \ref{equ:1} and integrate it into the training of our proposed \#maskUp algorithm.

\begin{equation}
\label{equ:1}
  {\cal L}(\theta) = {\cal L}_B(\theta) + \sum_i \frac{\lambda}{2} F_i (\theta_i - \theta^*_{A,i})^2
\end{equation}

where $ {\cal L_B}(\theta) $ is the loss for task B only, $\lambda$ defines how important the old task is compared to the new one and $i$ labels each parameter.

Continual learning has seen considerable applications within the NLP domain, such as for learning distinct tasks over changing languages \cite{castellucci-etal-2021-learning}, text classification employing regularization based on information regularization \cite{https://doi.org/10.48550/arxiv.2104.05489}, among others.

To the best of our knowledge, this is the first work adopting Continual Learning for Named Entity Recognition when training Transformer-based LLMs in an incremental number of tasks or over different distributions.

\section{Methodology}
\label{methodology}

\subsection{Problem Formulation}
\label{problem-formulation}

We introduce the notations about NER before getting into the details of the framework. For NER task, we denote ${(X_i, y_i)}_{i=1}^{N^s}$ as a training set with $N^s$ samples, where $X_i$ is the input text and $y_i$ is the NER label. Given a sentence with $m$ words, the input text can be formatted as $\textbf{X}=\{\textbf{x}_1, \textbf{x}_2,..., \textbf{x}_{m} \}$ and the NER label is $\textbf{Y}=\{\textbf{y}_1, \textbf{y}_2,..., \textbf{y}_{m} \}$. To solve the NER task, we try to maximize the posterior probability $p(\textbf{Y}|\textbf{X})$. From here, we can extract entities based on predicted $\textbf{y}_i$. With NER formulated, our aim shifts to defining objectives (1) Updating for Distribution Shifts (2) Learning over Distinct Tasks.

\subsubsection{Updating for Distribution Shifts}

Given an NER task, the data distribution may tend to shift gradually over time. This may occur due to variations in linguistic choices as time progresses or an entire shift in the application area. In such a case, where the task remains the same, i.e., identifying named entities belonging to some pre-selected categories, we consider the task to have undergone a distribution shift.

We formally interpret said objective as follows. Given a pre-selected list of categories $C=\{c_1, c_2,...c_{N^c}\}$, where $N^c$ is the total number of categories, one consolidates the entire model as a single entity, passing its weights entirely from one task to another. In this procedure, EWC is applied on the complete neural network, including both the transformer-based model as well as the output layer. In this case all tasks $T\_List = \{T_0, T_1,...,T_{N^t}\}$, where $N^t$ is the total number of tasks to be learnt, must contain the same categories for named entities.

\subsubsection{Learning Over Distinct Tasks}

On the contrary, to the above task, NERDA-Con also allows model dissociation and regularization solely over shared parameters. This interpretation of the standard EWC learning strategy will enable one to train NER models across different tasks. In such a case, the target classes may not belong to the same categories or may not be of the same length, i.e. the number of classes ($N_0^c$) for task 0 does not have to be equal to the total number of classes of task 1 ($N_1^c$). However, this framework works best between associated tasks, such as moving from a coarse-grained classification task to a fine-grained classification task.

\subsection{Implementation Details}

We build on both the objectives mentioned above, making sure NERDA-Con delivers for both distinct task learning as well as being robust to distribution shifts. For a given task $T_i$, we initialize a large pre-trained language model ($transformer\_model$) from the HuggingFace API. Once intialized we augment training data to create classification tasks for each entity in a given sentence $\textbf{X}$, where it can be represented in the form of words as, $\textbf{X}=\{\textbf{x}_1, \textbf{x}_2,..., \textbf{x}_{m} \}$ (for a sentence with $m$ words). We then define hyperparameters (1) EWC's $\lambda$, and (2) $shared\_params$ which would include only those parameters which are to regularized in the case of differing tasks (3) $output\_layer$ which incorporates $N^c+1$ classes, where $N^c$ is the number of classes defined in the given task and an extra class for default entities.

\begin{table*}[!ht]
    \centering
    \begin{tabular}{|l|l|l|l|l|}
    \hline
        ~ & \multicolumn{2}{l|}{\textbf{Task 1: CoNLL}} & \multicolumn{2}{l|}{\textbf{Task 2: DaNE}} \\ \hline
        \textbf{Level} & \textbf{NERDA-Con} & \textbf{Naive} & \textbf{NERDA-Con} & \textbf{Naive} \\ \hline
        B-PER & 0.8284 & 0.8828 & 0.8358 & 0.7102 \\ \hline
        I-PER & 0.8566 & 0.9143 & 0.9124 & 0.8244 \\ \hline
        B-ORG & 0.8156 & 0.7303 & 0.5973 & 0.5567\\ \hline
        I-ORG & 0.7638 & 0.6181 & 0.5625 & 0.2419  \\ \hline
        B-LOC & 0.7930 & 0.7905 & 0.7843 & 0.6556  \\ \hline
        I-LOC & 0.0833 & 0.1638 & 0.1022 & 0.1052 \\ \hline
        B-MISC & 0.5887 & 0.5531 & 0.4318 & 0.440 \\ \hline
        I-MISC & 0.5920 & 0.5422 & 0.5454 & 0.0975  \\ \hline
        AVG\_MICRO & \textbf{0.7761} & 0.7494  & \textbf{0.7145} & 0.5961 \\ \hline
    \end{tabular}
    \caption{Tabular comparison of F1-Score performance of proposed methodology and naive approach on the CoNLL dataset, after successive trainings over DaNE and Few-NERD datasets. Coupled with F1-Score performance of proposed methodology and naive approach on the DaNE dataset, after successive training over Few-NERD dataset, while incorporating EWC from previous training over the CoNLL dataset.}
    \label{tab:conll}
\end{table*}

\begin{table}[!ht]
    \centering
    \begin{tabular}{|l|l|l|}
    \hline
        \textbf{Level} & \textbf{NERDA-Con} & \textbf{Naive} \\ \hline
        art & 0.7827 & 0.7792 \\ \hline
        person & 0.9009 & 0.8987 \\ \hline
        event & 0.7156 & 0.7085 \\ \hline
        building & 0.7089 & 0.6911 \\ \hline
        location & 0.8275 & 0.8279 \\ \hline
        product & 0.6904 & 0.6843 \\ \hline
        other & 0.7282 & 0.7131 \\ \hline
        organization & 0.7580 & 0.7585 \\ \hline
        AVG\_MICRO & 0.7840 & \textbf{0.7882} \\ \hline
    \end{tabular}
    \caption{Tabular comparison of F1-Score performance of proposed methodology and naive approach on the Few-NERD dataset, while incorporating EWC from previous training over the CoNLL \& DaNE datasets.}
    \label{tab:few}
\end{table}

\section{Experiments}

\subsection{Datasets and Experimental Settings}

We categorize our experiments into two benchmarks based on the aforementioned objectives. 

\textbf{Distribution Shift:} We define distribution shift for multilingual models such that each task is represented by a different language with same output entity labels. In our case we use the following to represent each task:

\begin{itemize}
    \item CoNLL-2003 dataset \cite{conll-2003} (English)
    \item DaNE dataset \cite{hvingelby-etal-2020-dane} (Danish)
    \item Few-NERD dataset \cite{ding2021few} (Named Entities within English).
\end{itemize}

Task 1 and task 2 differ by language, whereas task 1 and task 3 differ by their respective class labels.  Both tasks 1 \& 2 categorize between the same labels display in Table~\ref{tab:conll}, whereas task 3 has a distinct set of labels presented in Table~\ref{tab:few}.

\textbf{Distinct Tasks:} We utilize coarse-grained data and fine-grained data from the Few-NERD dataset \cite{ding2021few}. The two tasks consist of (a) 8 coarse-grained types as mentioned above and (b) 66 fine-grained types such as $building-hotel$ and $location-park$ where-in they belong to one of the coarse-grained classes as well.

For the experiments, we used a Tesla P100 16GB as GPU, with 13GB RAM Intel Xeon as CPU. Note that each experiment is run three times, giving us 3-fold cross-validated results.

\subsection{Results}

\subsubsection{Distribution Shift}

We perform a progressive task-wise training routine to compute the performance of NERDA-Con against Naive approaches. For the formulation of naive approach, we assume model transformer weights as carried over from the previous task to the next, essentially making it a transfer learning task. We present the results of progressive learning for all three datasets (CoNLL, DaNE, and Few-NERD) in Tables~\ref{tab:conll}-\ref{tab:few}. 

Even though CoNLL was in English (same as Few-NERD dataset) and incorporated the same labels as the DaNE dataset, it was still able to retain a higher performance throughout successive tasks. It improved upon the $AVG\_MICRO$ F1-Score of Naive approach by 2.67\%. On the other hand, a complete language shift was significantly impacted during successive training on an English task. Even so, NERDA-Con was able to retain a better performance of 0.7145 $AVG\_MICRO$ F1-Score compared to Naive's 0.5961. Finally, for the third task, Few-NERD dataset, we can see that there's effectively not much difference (-0.0042 of NERDA-Con's $AVG\_MICRO$ F1-Score from the Naive approach) allowing us to experimentally validate that NERDA-Con allows learning new tasks, irrespective of previous experience of differing tasks.

% \begin{table}[!ht]
%     \centering
%     \begin{tabular}{|l|l|l|}
%     \hline
%         \textbf{Level} & \textbf{NERDA-Con} & \textbf{Naive} \\ \hline
%         B-PER & 0.8358 & 0.7102 \\ \hline
%         I-PER & 0.9124 & 0.8244 \\ \hline
%         B-ORG & 0.5973 & 0.5567 \\ \hline
%         I-ORG & 0.5625 & 0.2419 \\ \hline
%         B-LOC & 0.7843 & 0.6556 \\ \hline
%         I-LOC & 0.1022 & 0.1052 \\ \hline
%         B-MISC & 0.4318 & 0.440 \\ \hline
%         I-MISC & 0.5454 & 0.0975 \\ \hline
%         AVG\_MICRO & \textbf{0.7145} & 0.5961 \\ \hline
%     \end{tabular}
%     \caption{Tabular comparison of F1-Score performance of proposed methodology and naive approach on the DaNE dataset, after successive training over Few-NERD dataset, while incorporating EWC from previous training over the CoNLL dataset.}
%     \label{tab:dane}
% \end{table}

\subsubsection{Distinct Tasks}

\begin{figure}
    \centering
    \includegraphics[width=\columnwidth,height=5.5cm]{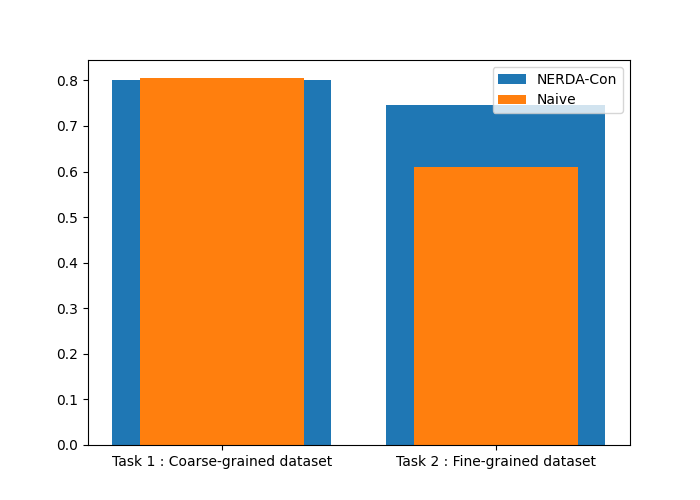}
    \caption{Performance evaluation for Distinct Tasks on the Few-NERD dataset. Task 1 : training on coarse-grained labels and Task 2 : training on the fine-grained dataset.}
    \label{fig:distinct}
\end{figure}

A clear distinction can be seen when validating experiments belonging to completely different tasks. As shown in Figure~\ref{fig:distinct} NERDA-Con is able to retain information over the $transformer\_model$. An improvement of 0.1366 in F1-Score is achieved.

The results strongly affirm that NERDA-Con allows successive learning over both (1) distribution shifts, as well as (2) distinct tasks, allowing us to satisfy both the objectives defined in Section~\ref{problem-formulation}. Our results substantiate our claims for NERDA-Con, a library extending LLM based NERs with continual learning.

\section{Conclusion}

With an aim to update over distribution shifts and learn individually separate tasks, we present NERDA-Con, a first of its kind open-sourced pipeline which extends continual learning over large language models based NER systems. By providing  3-fold cross-validated results, we describe how NERDA-Con can successfully learn over three different datasets, which differ by language as well as class labels and can significantly learn and adapt from all data. Since NER models are often the first step in the construction of knowledge bases as well as the identification of certain information, we believe NERDA-Con will be able to save a significant amount of memory and time computationally. We acknowledge that our NER pipeline may be susceptible to traditional biases plaguing other NER systems\cite{ghaddar-etal-2021-context,https://doi.org/10.48550/arxiv.2008.03415}, which we aim to address in future work.

\bibliography{example_paper}
\bibliographystyle{icml2022}

% \newpage
% \appendix
% \onecolumn
% \section{Detailed Performance Results for Distribution Shifts}

% \begin{figure}
%     \centering
%     \includegraphics[width=1.1\textwidth]{imgs/conll.png}
%     \caption{Performance of NERDA-Con on the CoNLL dataset after training over all 3 tasks, compared against the Naive approach. There can be seen a reduction in performance between the Naive approach when compared to our proposed methodology.}
%     \label{fig:conll}
% \end{figure}

% \begin{figure}
%     \centering
%     \includegraphics[width=1.1\textwidth]{imgs/dane.png}
%     \caption{Performance of NERDA-Con on the DaNE dataset after training over the remaining two tasks, compared against the Naive approach. There can be seen a drastic improvement in F1-Score between our proposed methodology and the Naive approach.}
%     \label{fig:dane}
% \end{figure}

% \begin{figure}
%     \centering
%     \includegraphics[width=1.1\textwidth]{imgs/ner.png}
%     \caption{Performance of NERDA-Con on the Few-NERD dataset after training over itself, compared against the Naive approach. There can be seen little to no difference in F1-Score between our proposed methodology and the Naive approach, ensuring that our model remains  considerably trainable with EWC.}
%     \label{fig:few}
% \end{figure}

\end{document}